\documentclass[a4paper,twoside]{article}

\usepackage{epsfig}
\usepackage{subfigure}
\usepackage{calc}
\usepackage{amssymb}
\usepackage{amstext}
\usepackage{amsmath}
\usepackage{amsthm}
\usepackage{multicol}
\usepackage{pslatex}
\usepackage{apalike}
\usepackage{SCITEPRESS}     

\usepackage{csquotes} 
\usepackage{float}
\usepackage{booktabs}
\usepackage[table,xcdraw]{xcolor}

\newcommand{\leqnomode}{\tagsleft@true\let\veqno\@@leqno}
\usepackage{cuted}

\begin{document}

\title{Spontaneous Facial Expression Recognition using Sparse Representation}

\author{\authorname{Dawood Al Chanti\sup{1} and Alice Caplier\sup{1}}
\affiliation{\sup{1}Univ. Grenoble Alpes, GIPSA-Lab, F-38000 Grenoble, France CNRS, GIPSA-Lab, F-38000 Grenoble, France}
\email{dawood.alchanti@gmail.com}}

\keywords{Dictionary learning, Random projection, Spontaneous facial expression, Sparse representation.}

\abstract{Facial expression is the most natural means for human beings to communicate their emotions. Most facial expression analysis studies consider the case of acted expressions. Spontaneous facial expression recognition is significantly more challenging since each person has a different way to react to a given emotion. We consider the problem of recognizing spontaneous facial expression by learning discriminative dictionaries for sparse representation. Facial images are represented as a sparse linear combination of prototype atoms via Orthogonal Matching Pursuit algorithm. Sparse codes are then used to train an SVM classifier dedicated to the recognition task. The dictionary that sparsifies the facial images (feature points with the same class labels should have similar sparse codes) is crucial for robust classification. Learning sparsifying dictionaries heavily relies on the initialization process of the dictionary. To improve the performance of dictionaries, a random face feature descriptor based on the Random Projection concept is developed. The effectiveness of the proposed method is evaluated through several experiments on the spontaneous facial expressions DynEmo database. It is also estimated on the well-known acted facial expressions JAFFE database for a purpose of comparison with state-of-the-art methods.}

\onecolumn \maketitle \normalsize \vfill

\section{\uppercase{Introduction}}
\label{sec:introduction}

\noindent An increasing number of techniques have been proposed in the literature for emotional facial expressions analysis since emotion is an essential component of interpersonal relationships and communication. Human behavior is central to research concerning interaction processes. A facial image contains much information about a person identity but also about emotion and state of mind. Emotion cues show how we feel about ourselves and others. The cues are represented through facial components (eyes, nose, mouth, cheeks, eyebrow,forehead, etc) which are the region of interest (ROI) for emotional recognition system. Facial expression recognition system utilized in locating and extracting different facial motions and facial feature changes from the ROI region and classifying  into one of the emotional or mental states. The potential utility of a system capable of analyzing spontaneous facial expressions automatically is considerable in terms of its potential applications: human machine interaction, detection of mental disorders, remote detection of people in trouble, detection of malicious behavior and multimedia facial queries \cite{3}. The current researches about facial expression recognition can be divided into two categories \cite{peng2009research}: recognition of facial affect and recognition of facial muscle actions. In this paper, facial affect recognition is considered for observable expressions of emotion displayed through facial expressions. Our choice comes from the fact that it provides simplicity in identification of the various emotions via extracting information about facial expressions from images. However, facial action units (AUs) which are related to the contraction of specific facial muscles, consist of 44 action units. Although the number of atomic action units is small, more than 7,000 combinations of action units have been observed \cite{scherer1982handbook}.

As far as automatic facial affect recognition is concerned, most of the existing efforts focused on the six basic Ekman's-emotions \cite{1} because those emotions have universal properties. Moreover, relevant training and test materials are available (e.g., \cite{25} and \cite{27}). These studies are limited to exaggerated expressions and controlled environments. There are a few tentatives efforts to detect non-basic affective states including mental states (\enquote{irritated}, \enquote{worried}...) \cite{26}. But those expressions are closer to natural behavior. Additionally, the fact is that spontaneous facial expressions have different temporal and morphological characteristics than posed ones. 

The purpose of our work is to demonstrate that sparse representation is an efficient model in order to classify and to increase the accuracy rate of predicting the spontaneous facial expressions using spontaneous facial images. Sparse representation provides higher or lower dimensional representations which induce the likelihood that image classes will be possibly linearly separable. The sparse discriminative feature set provides the main interface through which a machine learning algorithm can infer about the data. More precisely, the main issue with sparse representation being dictionary learning and due to the fact that the original facial image has a very high dimension, the straightforward application of sparse representation for sparse feature extraction from raw images does not lead to a meaningful sparse representation. Thus we present an efficient initialization strategy and dimensionality reduction technique via developing an optimized random face feature descriptor (RFFD) based on the random projection (RP) concept \cite{49}. RFFD aims at projecting the facial images into a lower dimensional space and at selecting the most discriminative feature sets that minimizes the correlation between different facial image classes while maximizing the correlation within facial image classes, in an attempt to ensure the uniqueness of the atoms selection from the dictionary during sparse coding process. Our pre-training step allows us to avoid high computational resources (memory usage and training time) required during dictionary training which is an important requirement for developing a real-time automatic facial expression recognition system. Experimental results on the JAFFE acted facial expression database and on the DynEmo spontaneous expression database demonstrate that our algorithm outperforms many recently proposed sparse representation and dictionary learning based approaches. Our algorithm has the capacity to be trained on a small or a big dataset and to provide a high accuracy rate, which can be considered as an advantage compared to deep learning approaches which are doing great nowadays only if a big dataset is provided.

\section{\uppercase{Related Work}}
\label{sec:relatedwork}

\noindent Numerous methods for extracting discriminative information about facial expressions from images have been developed. For example, Eigenfaces, Fisherfaces, and Laplacianfaces have been used on full face images \cite{36}. Gabor filter banks also have been successfully used as an efficient facial feature (\cite{31} and \cite{32}) because these features are locally concentrated and have been shown to be robust to block occlusion \cite{donoho2006compressed}. Once the feature vector is extracted from an image, this vector feeds a classifier which gives the recognized expression. A survey of automatic facial expression recognition methods is presented in \cite{35}.

A noteworthy contribution of sparse representations of signals has been reported in recent years. It has been successfully applied to a variety of problems in computer vision and image analysis, including image denoising \cite{17}, image restoration \cite{20} and image classification \cite{24}, \cite{37} and \cite{16}. Sparse representation modeling of data assumes an ability to describe signals as linear combinations of few atoms from a pre-specified dictionary. The success of the model relies on the quality of the dictionary that sparsifies the signals. The choice of a proper dictionary can be done using one of two following ways \cite{14}: building a sparsifying dictionary based on a mathematical model of the data (wavelets, wavelet packets, contourlets, and curvelets), or learning a dictionary to perform best on a training set. Reference \cite{37} employs the entire set of training samples as the dictionary for discriminative sparse coding, and achieves impressive performance for face recognition. Many algorithms (\cite{19} and \cite{22}) have been proposed to efficiently learn an over-complete dictionary (the number of prototype signals, referred as atoms, is much greater than the features size) that enforces some discriminative criteria. Recently, another sparse representation for object representation and recognition was proposed in the seminal work \cite{37}. In \cite{29}, the class labels of training data are used to learn a discriminative dictionary for sparse coding. In addition, label information is associated with each dictionary item to enforce discriminability in sparse codes during the dictionary learning process. More specifically, a new label consistency constraint called \enquote{discriminative sparse-code error} is introduced and combined with the reconstruction error and the classification error to form a unified objective function.

Our work is inspired by the good reputation of sparse representation in both theoretical research and practical applications (\cite{24}, \cite{37}, \cite{16} and \cite{20}). Moreover, our choice comes from the fact that sparse representation has the ability to provide sparse vectors that can share the same sparsity pattern at class level if it is correctly built.

\section{\uppercase{Model Architecture}}
\label{modelarichitecture}

\begin{figure}[!h]
	\centering
	\includegraphics[width=1.8 in]{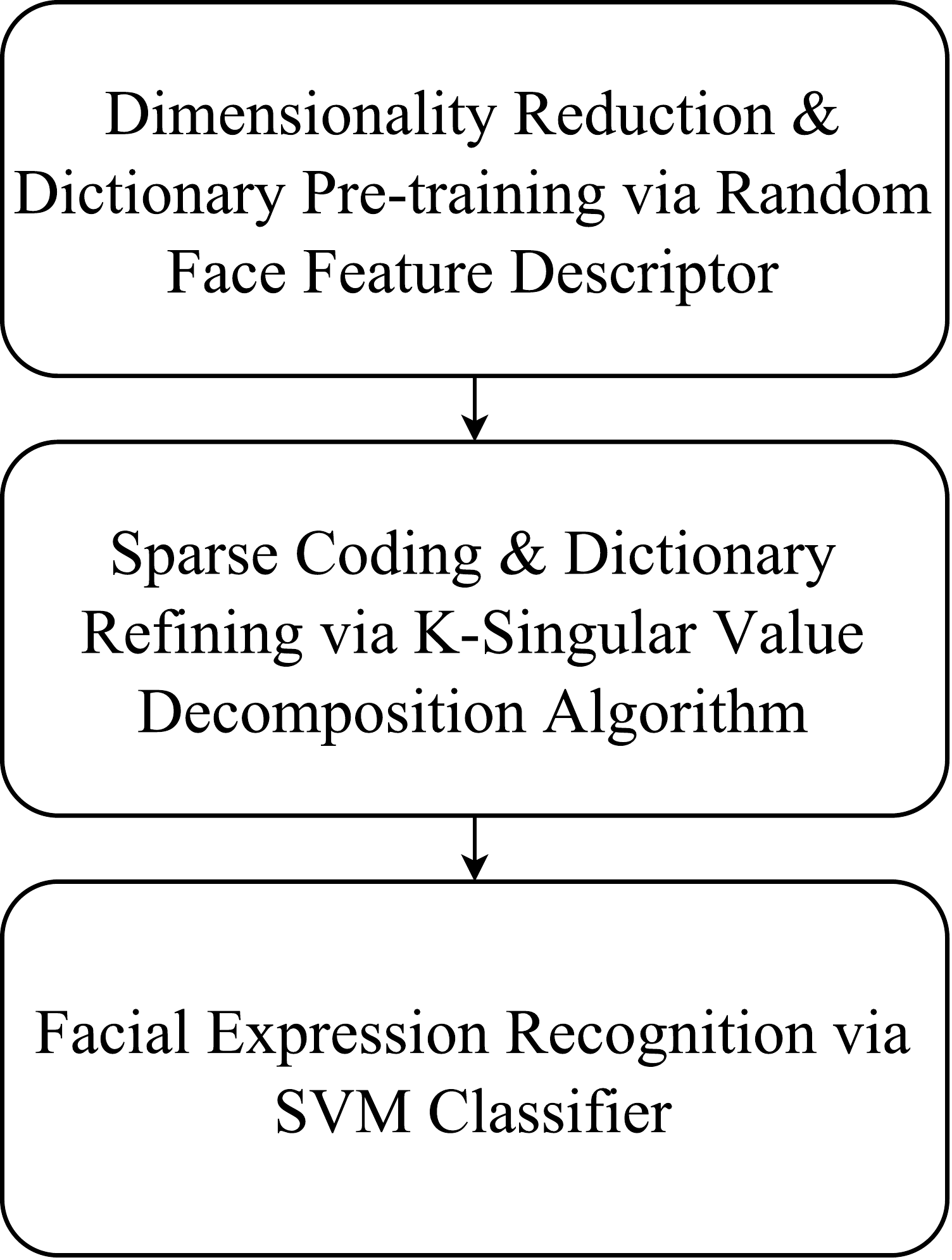}
	\caption{Global architecture of spontaneous facial expression recognition algorithm (SFER).}
	\label{fig:GMA}
\end{figure}

\noindent Figure \ref{fig:GMA} presents the global architecture of the proposed algorithm for facial expression recognition.

\subsection{Dimensionality Reduction and Dictionary Pre-training Stage}
\label{DPT}
\noindent We aim to leverage the random projection (RP) technique \cite{39} to develop a random face feature descriptor (RFFD) for dictionary pre-training that elegantly solves the problem of shared subspace distribution. It also projects the raw data into a lower-dimensional space, while preserving their reconstructive and discriminative properties. Beside, it seeks for the best transformation matrix that maximizes the separation between the multiple classes which is the main key to induce sparsity. 

As a pre-processing stage, a face detector \cite{4} is applied in order to detect and locate a bounding box around the face. Facial images are cropped to focus on the expressive parts only: eyes, eye-brows, mouth and nose, in order to reduce the effect of background variation. Then, RFFD is applied in order to project the data into a lower dimensional subspace and to extract the most informative and discriminative independent features. The projected data serve as dictionary initialization. Thus, the dictionary is pre-trained with  feature vectors sharing same patterns within class label while feature vectors from different classes have different patterns.

\vfill
\noindent \textbf{Random Projection Theory and Concept}
\vfill
Random projection is a powerful data dimension reduction technique because it is capable to preserve the reconstructive properties of the data \cite{39}. It uses random projection matrices whose columns have unit length to project data from high-dimensional subspace to a low-dimensional subspace. It is a computationally simple and efficient method that preserves the structure of the data without significant distortion \cite{9}. 

The concept of RP is as follows: Given a data matrix $X$, the dimensionality of data can be reduced by projecting it onto a lower-dimensional subspace formed by a set of random vectors:
\begin{equation} 
    A^{m \times N} = R^{m\times d} \cdot X^{d \times N}
\end{equation}

where $N$ is the total number of points, $d$ is the original dimension, and $m$ is the desired lower dimension, $R$ is the random transformation matrix and $A$ is the projected data. The central idea of RP is based on the Johnson-Lindenstrauss lemma. For complete proofs of the lemma refer to \cite{47}. 

The choice of the random matrix $R$ is one of the crucial points of interest. Reference \cite{47} employs a random matrix $R$ whose elements are drawn independently and are identically distributed (i.i.d.) from a zero mean, bounded variance distribution. There are many choices for the random matrix. A random matrix with elements generated by a normal distribution $r{_{i,j}} \sim$ $N(0,1)$ being one of the simplest in terms of analysis \cite{47} has been chosen in this work.
\vfill
\noindent \textbf{Random Face Feature Descriptor Algorithm:}
\vfill

A Random Face Feature Descriptor based on RP concept is designed. RFFD firstly tackles the curse of dimensionality in which each image is projected onto $m$-dimensional vector with a randomly generated projection matrix $R$ from a zero-mean normal distribution. Each row of the transformation random matrix is $l_{2}$ normalized. RFFD aims at minimizing the correlation between different classes while maximizing the correlation within-classes. It preserves discriminative properties of the input data. 

\begin{small}
\begin{figure}[]
\centering
\framebox{
\begin{minipage}{\dimexpr\linewidth-2\fboxrule-2\fboxsep}
\begin{small}

\end{small}

\begin{small}
\textit{Input:}
\begin{itemize}
   \item $X \in R^{d \times N}$ : is the input matrix, where each column represents one sample. $d$ is the original dimension of the image and $N$ is the total number of samples. 
    \item $M = [m{_{1}},m{_{2}},...,m{_{dd}}]$: a list of possible desired lower dimensions.
    \item $rn$: is the desired number for generating good random matrices.
\end{itemize}
\end{small}

\begin{small}
\textit{Algorithm:}

\begin{enumerate}
    \item For each $m$ in $M$
    
    \begin{enumerate}
     \item While $j$ in $rn$
                \begin{enumerate}
                   \item Generate random matrix $R^{[m \times d]}_{j}$ from zero mean, normal distribution $N(0,1)$ and $l_{2}$ normalized columns.
                     \item Compute $A_{[m \times N]} = R^{[m \times d]}_{j} \cdot X_{[d \times N]}$.
                     \item Check the quality of the obtained features in $A_{[m \times N]}$.
                    \item If $A_{[m \times N]}$ has good quality features, add $A_{[m \times N]}$  to a list $L_{m}$
                    \end{enumerate}  
    
     \item For each $A$ in $L_{m}$
     
                     \begin{enumerate}
                            \item Apply Linear SVM over the obtained $(A)$ with cross validation.
                            \item Store the recognition rate.
                    \end{enumerate}  

     \item Pick up the best $A$ among $A$'s in $L_{m}$ that reaches the highest classification accuracy rate.
     \item Add best $A$ and its classification accuracy rate to a list $L_{mR}$
    \end{enumerate}

    \item Pick up in the $L_{mR}$ list the $A$ that reaches the highest accuracy rate and the corresponding best transformation matrix $R_{j}$.
    
\end{enumerate}
\end{small}

\begin{small}
\textit{Output:}
\begin{itemize}
   \item Projected data $A_{[m \times N]}$ from the optimal $R$ and $m$.
    \item Optimal projection matrix $R$ that generates the best discriminative features.
    \item Optimal lower dimension $m$.
\end{itemize} 

\end{small}
\end{minipage}
}
\caption{Random Face Feature Descriptor Algorithm.}
\label{fig:RFFD}
\end{figure}
\end{small}

Figure \ref{fig:RFFD} presents the RFFD algorithm. It looks for the best projection matrix $R$ and the best dimension of projection $m$ that preserve the structure and the reconstructive properties of the original data. The intuition behind this algorithm is as follows: since $R$ is generated randomly, it is not guaranteed to have good quality features. A good quality feature vector $A_{i}$, $i=[1,...,m]$, is a vector where its most entire elements are not full of zeros. The quality of the projected matrix $A^{m \times N}$ (figure \ref{fig:RFFD} step iii) is checked by thresholding the norm of every column vector $A_{i}$. If the norm of $A_{i}$ is smaller than a given threshold, it is considered as a bad feature vector. 

Once a good data projection $A^{(m,N)}$ is obtained, $R$ is considered as good random transformation matrix. Moreover, the quality of the features vectors $A_{i}$ from two different good transformation matrices $R$ and $R^{\prime}$ can vary. We aim at picking out $R$ that induce the most discriminability between classes. Selecting the best $R$ among a set of $R$'s is then important (figure \ref{fig:RFFD} steps b and c). In addition, selecting the best dimension $m$ that preserves the discriminative properties of the original data with minimal distortion has a great effect on the final recognition rate (figure \ref{fig:RFFD} steps d and 2). 

The projected data obtained as the output of the RFFD process is used to initialize the dictionary that is required for the sparse representation process. This step is important to induce sparsity during learning process by initializing the dictionary with atoms that are highly informative and that have maximum separation between multiple classes.

\subsection{Dictionary Refining and Sparse Coding Stage} 
\label{DRSC}

\noindent The second step of the algorithm (see figure \ref{fig:GMA}) firstly aims at refining the pre-trained dictionary to sparsify the images via K-Singular Value Decomposition (K-SVD) algorithm \cite{52}. Secondly it aims at deriving the sparse code associated to each signal by solving $l_{0}$-norm regularization to enforce sparsity by using an approximate sparse reconstruction algorithm, Orthogonal Matching Pursuit (OMP) \cite{tropp2004greed}.
\\
Given a dataset $Y \in R^{n \times N}$ and a target sparsity level L (maximum number of atoms allowed in each representation) the problem is to build the dictionary $D \in R^{n \times K}$ and the sparse matrix $X \in R^{K \times N}$ such that $Y \approx DX$. The problem can be formulated as:

\begin{equation} 
<D,X> = \underset{D,X}{\operatornamewithlimits{argmin}} ||Y - DX||_{F}^{2}\:\:\:\:\:\: \textnormal{such that} \:\:\:\:\:\: 
\label{eq0}
\end{equation}

\[\forall_{i}, i \in [1,N], ||x_{i}||_{0} \leq L \]
\[\forall_{j}, j \in [1,K], ||d_{j}||_{2}=1
\]

where:
\begin{itemize}
\item $||x_{i}||_{0}$ is $||l||_{0}$ pseudo norm, defined by  the number of non-zero coefficients in column $x_{i}$.
\item $||E||^{2}_{F}$ is the Frobenius norm.
\item columns $d_{j}$ is $l_{2}$ normalized atom of the dictionary $D$.
\end{itemize}

Equation \ref{eq0} can be solved by an alternating two step optimization process :

\begin{enumerate}
\item \textbf{Sparse Coding}: Keep the pre-trained dictionary $D$ fixed and estimate $X$ such as: 

\begin{align}
\begin{split}
   X = \underset{X}{\operatornamewithlimits{argmin}}  ||Y - DX||_{2}^{2} \\
  \textnormal{such that} \:\:\: \forall_{i}, ||x_{i}||_{0} \leq L
\end{split}
\label{eq1}
\end{align}

The sparse representation $X$ is optimized by using the OMP algorithm. Compared with other alternative methods for sparse coding, a major advantage of the OMP is its simplicity and fast implementation.

\item \textbf{Dictionary Refining}: Keep the obtained sparse matrix $X$ fixed and update the pre-trained dictionary $D$ via K-SVD algorithm to better fit the data.
\end{enumerate}

To recap, the search for the sparse representation of facial expression images over a pre-trained dictionary is achieved by optimizing an objective function (equation \ref{eq0}) that includes two terms: one that measures the signal reconstruction error and the other that measures the best sparsity level to ensure the correct representation of the signals. 

\subsection{Classification Stage}
\label{CLS}
 
\noindent In the last step (see figure \ref{fig:GMA}), the sparse matrix $X$ is directly used as feature vectors for classification. Our model trains a \enquote{Multinomial Linear Support Vector Machine} classifier \cite{15} for the purpose of facial expression recognition. We consider linear SVM classifier among the others well-known classifiers (i.e., K-Means, Ada Boost and Decision Tree) since it shows the best results. In the training step, the sparse matrix $X_{training\_data}$ is used to learn a predictive model to recognize facial expressions. The test sparse matrix $X_{test\_data}$ is used for generalization purpose: the capability of the model to predict unseen facial expression is tested. Grid search is applied to find the best parameter $C$ (regularization parameter) to tune the linear SVM classifier.

\section{Experimental Setup and Analysis} 
\label{experimentalsetup}

\noindent A critical experimental evaluation of the proposed approach is presented. Two public data sets that exhibit various emotions in different conditions, starting from acted facial expressions: the JAFFE database \cite{27}, to everyday natural and spontaneous facial expressions: the DynEmo database \cite{28}, are used. The effectiveness of the proposed random face feature descriptor as a dimension reduction technique and dictionary pre-training method is analyzed by considering first the acted and controlled JAFFE database. This database is also used for fair comparison with the state of the art methods. Then experiments on the DynEmo database are reported since spontaneous facial expressions recognition is our main goal.

\subsection{Model Validation over the JAFFE Database}
\label{validation}

\begin{figure}[]
\centering
\includegraphics[width=3. in]{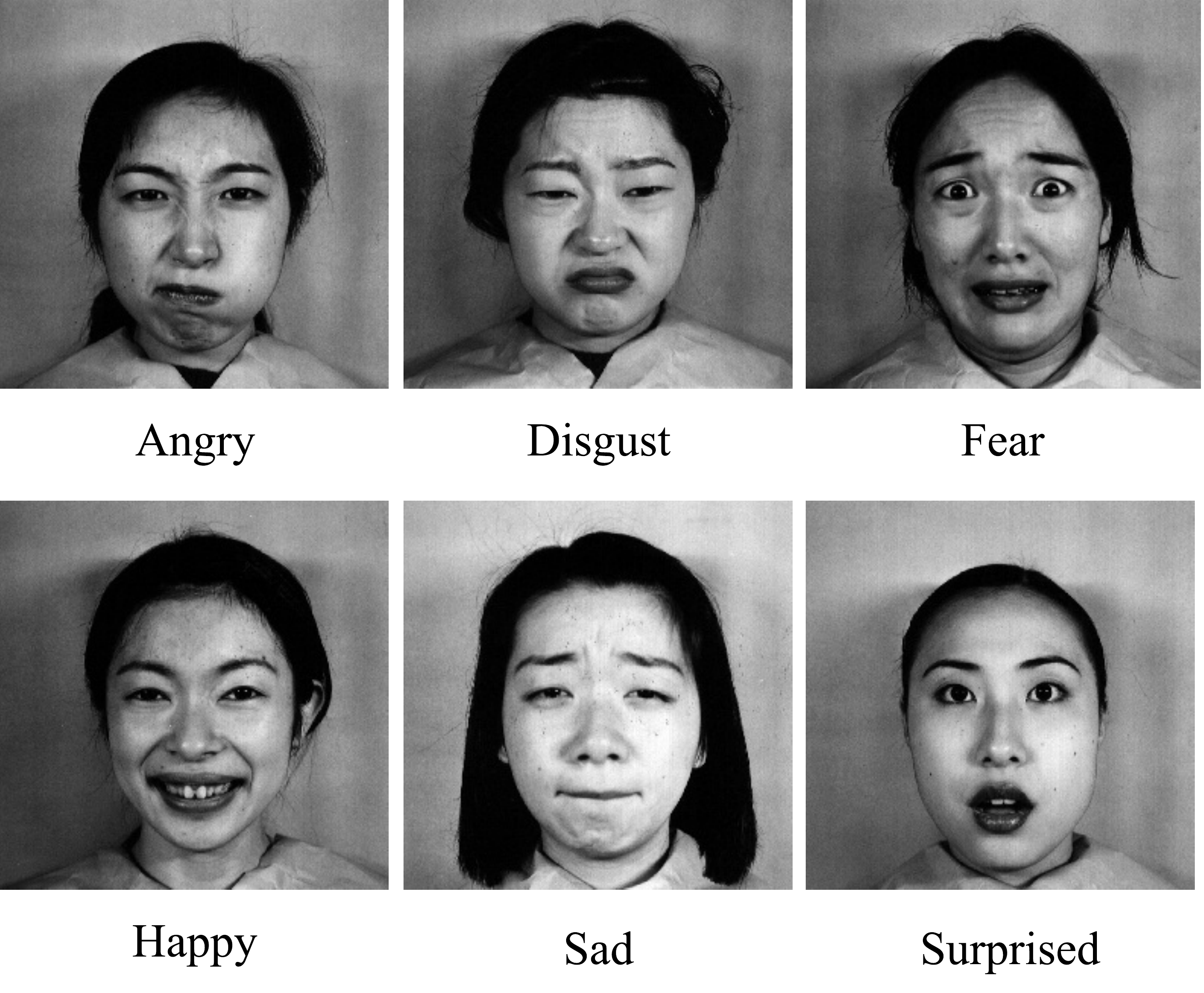}
\caption{Examples of facial expressions from JAFFE database.}
\label{fig:JAFFE}
\end{figure}

\textbf{The JAFFE Database:}
\noindent The Japanese Female Facial Expression (JAFFE) database is a well-known database made of acted facial expressions related to Eckman's emotions. It contains 213 images of female facial expressions including: \enquote{happy}, \enquote{anger}, \enquote{sadness}, \enquote{surprise}, \enquote{disgust}, \enquote{fear} and \enquote{neutral}. Resolution of original facial images is $256 \times 256$ pixels. After cropping, each image has a resolution of $138 \times 128$ pixels. The head is almost in frontal pose. The number of images corresponding to each of the seven categories of expressions is roughly the same (around 30 images per class). A few of them are shown in figure \ref{fig:JAFFE}. It is obvious that expressions are over exaggerated. Nonetheless this database has often been used in literature to evaluate the performance of some facial expressions recognition algorithms.

\vspace{3mm}

\noindent \textbf{JAFFE Database Protocol:}
Identities that appear in the training data sets do not appear in the test set.
\begin{enumerate}
     \item \textbf{train set}: 20 images per class are picked out as training set. In total we have 143 facial expression images, randomly shuffled, for training our algorithm.
     \item \textbf{development set}: Leave-one-out cross validation is considered over the training set to tune the algorithm parameters.
     \item \textbf{test set}: 10 images per class are picked out as test set. In total we have 70 facial expression images, randomly shuffled, to test the performance of our algorithm.
\end{enumerate}

\noindent \textbf{Experimental Setup and Analysis: }\\
The efficiency of our approach and its capability to recognize acted facial expressions beforehand testing it on spontaneous facial expressions is evaluated and demonstrated as a control experiment.

\begin{figure*}[ht]
\centering 
\includegraphics[width=5. in]{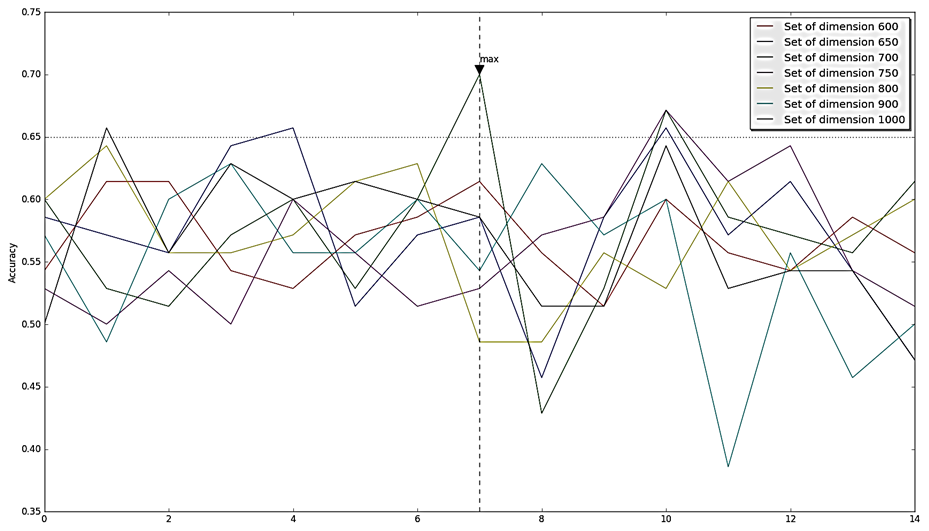}
\caption{Random Face Feature Descriptor Evaluation over the JAFFE Database.}
\label{fig:RFFDJAFFE}
\end{figure*}

\textbf{Firstly}, the dataset is divided into two portions based on the number of images per class as defined in the previous JAFFE dataset protocol. Figure \ref{fig:RFFDJAFFE} represents the evaluation of the \textit{random face feature descriptor}. The x-axis represents the generation of different random matrix $R$ for the same desired dimension $m$. The y-axis represents the final average classification rate over the projected data. To evaluate the performance of RFFD over the JAFFE database, we define a list of desired lower dimensions, $m$: $600, 650, 700, 750, 800, 900, and 1000$. For each dimension, different random matrices $R_{i}$ are generated. For a given dimension, $R$ that reaches the maximum average classification accuracy rate is picked out. Finally, both the best dimension and the best R are derived. Figure \ref{fig:RFFDJAFFE} shows that the optimal random projection matrix is $R^{700}_7$. Which means, the optimal dimension is found to be $700$ at random matrix $R_{7}$ of this set. The projected data from $R^{700}_7$ reaches an average classification rate of $70\%$.

We compare the proposed dimension reduction method with PCA which is probably the most popular method for dimensionality reduction. Our method out performs PCA method as shown in table \ref{RFFDPER}.

\begin{table}[]
\centering
\caption{RFFD versus PCA Performance on the JAFFE Database.}
\begin{tabular}{|c|c|}
\hline
\multicolumn{2}{|c|}{{\color[HTML]{000000} JAFFE database}} \\ \hline
{\color[HTML]{000000} \textbf{Projection Method}} & {\color[HTML]{000000} \textbf{Average recognition rate} \%} \\ \hline
{\color[HTML]{000000} RFFD (ours)} & {\color[HTML]{000000} 70} \\ \hline
{\color[HTML]{000000} PCA} & {\color[HTML]{000000} 30} \\ \hline
\end{tabular}
\label{RFFDPER}
\end{table}

\begin{figure*}[ht]
\centering
\includegraphics[width=5. in]{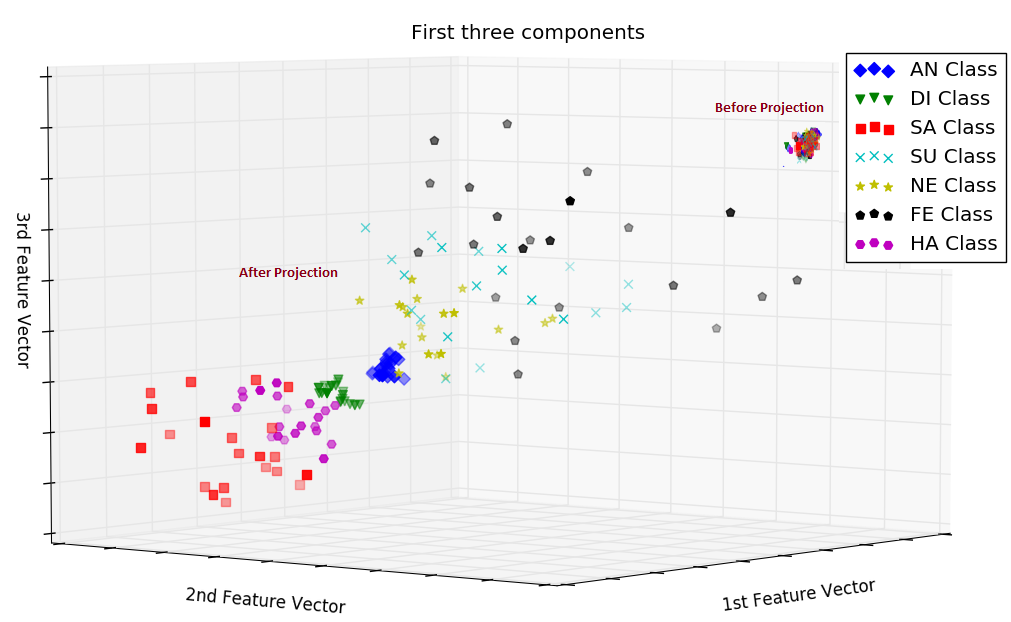}
\caption{Data distribution before and after projection.}
\label{fig:beforeRpJaffe}
\end{figure*}

For illustration, the first three feature vectors for 20 images per class before and after RFFD over the test data are displayed. Figure \ref{fig:beforeRpJaffe} shows that the data have a shared subspace before projection. This problem is solved by the proposed RFFD method since the data are then partially linearly separable. This reaches our main concern to obtain highly informative and independent feature vectors between different classes.

\begin{figure}[]
\centering
\includegraphics[width=3. in]{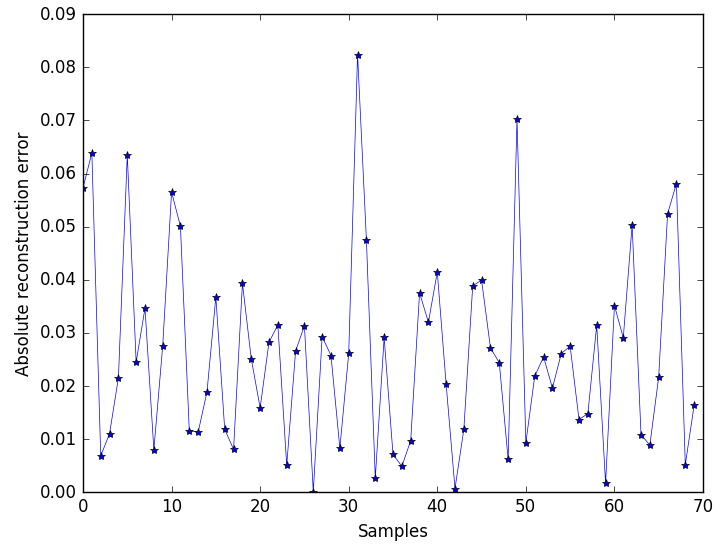}
\caption{Absolute reconstruction error for test samples.}
\label{fig:ErrorJaffe}
\end{figure}

\begin{figure*}[ht]
\centering 
\includegraphics[width=5. in]{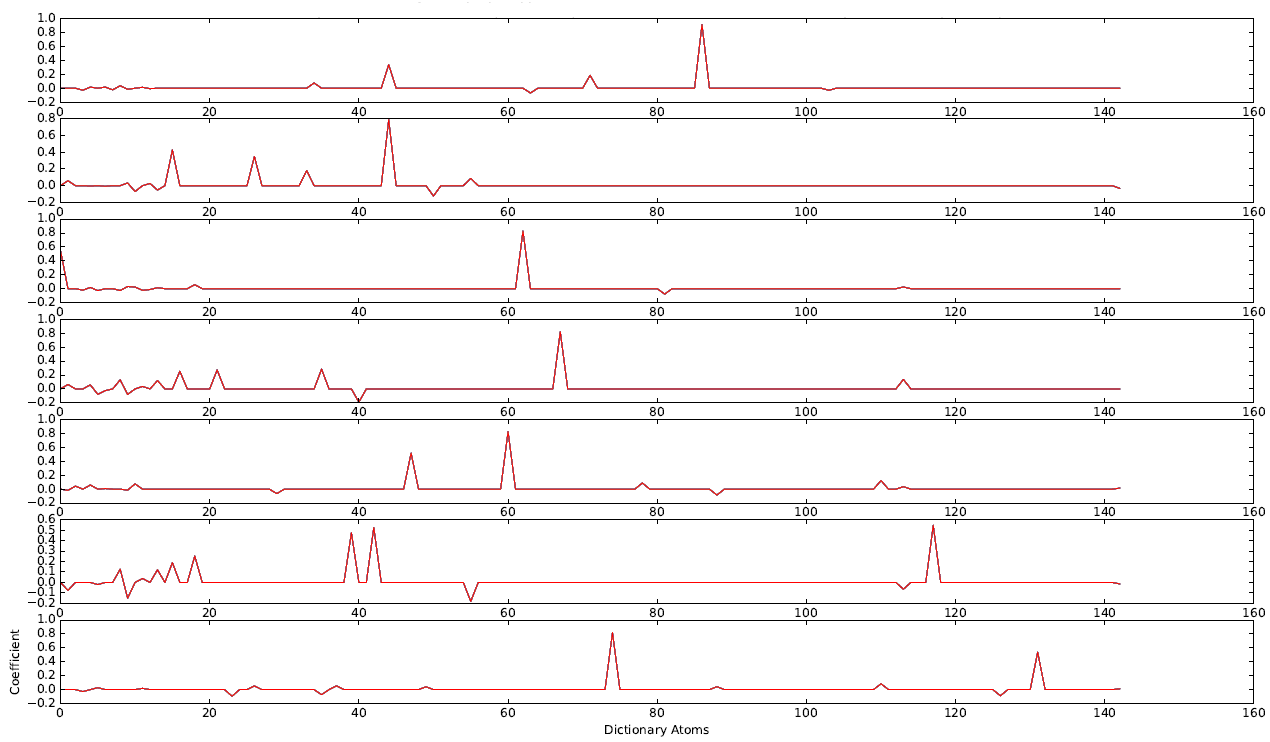}
\caption{Sparse code coefficients of signals representing AN, DI, FE, HA, SA, SU and NE expressions respectively from top to bottom.}
\label{fig:SPCORF}
\end{figure*}

\textbf{Secondly}, the optimal data projection is used to initialize the dictionary $D^{(0)}$, of size (700 features, 143 atoms ($K$)) (under-complete dictionary: number of the atoms is smaller than the feature size). Each column of the dictionary is normalized to have unit norm, which ensures that the angle is proportional to the inner product. K-SVD algorithm is applied to refine the initialized dictionary and the sparse matrix is computed via OMP algorithm. The optimal dictionary is obtained. It yields to a signal representation with the smallest possible support while the estimated signal is still close to the observation. The choice of $L$ is estimated to $15\%$ of the dictionary size  by controlling the absolute reconstruction error (figure \ref{fig:ErrorJaffe}) and the discriminability of the obtained sparse code (figure \ref{fig:SPCORF}). Figure \ref{fig:ErrorJaffe} shows the ability of the trained dictionary to reconstruct the test samples with minimal reconstruction error and with low sparsity level (21 non-zero coefficient at maximum). Figure \ref{fig:SPCORF}  represents the sparse code coefficients of a given image of the following expressions: "Anger", "Disgust", "Fear", "Happy", "Sad", "Surprise" and "Neutral" respectively from top to bottom. The x-axis represents the dictionary atoms (basis vectors) in which the coming facial image is encoded from while the y-axis represents the coefficients value. Figure \ref{fig:SPCORF} shows that each expression is encoded by a different set of atoms with different weights coefficients. The discriminability of the sparse code is a very important property for robust classification. The other point to be noticed is that under-complete dictionary allows faster computation, since OMP algorithm will pick out at most $L$ out of $K$ atoms (greedy algorithm).

\textbf{Finally}, after deriving the test and the train sparsity matrices via OMP algorithm based on the refined dictionary via K-SVD algorithm, a linear SVM classifier is trained over the training sparse matrix ($143$: samples, $143$: sparse feature vector size). The test sparsity matrix ($70,143$) is used to assess the ability of the classifier to generalize. A grid search is applied to find the best regularization parameter $C$, and $C$ is found to be $1$.

\begin{table}[]
\centering
\caption{Recognition Rate per Class \% on the JAFFE Database.}
\label{tabel1}
\begin{tabular}{|c|c|}
\hline
{\color[HTML]{000000} \textbf{Class}} & {\color[HTML]{000000} \textbf{Recognition Rate \%}} \\ \hline
AN & 99 \\ \hline
DI & 90 \\ \hline
FE & 95 \\ \hline
{\color[HTML]{000000} HA} & {\color[HTML]{000000} 89} \\ \hline
{\color[HTML]{000000} SA} & {\color[HTML]{000000} 100} \\ \hline
SU & 100 \\ \hline
NE & 91 \\ \hline
\end{tabular}
\end{table}

Table \ref{tabel1} presents the recognition rate per class. It shows that expressions \enquote{anger} (AN), \enquote{sadness} (SA) and \enquote{surprise} (SU), are perfectly classified. For the expressions \enquote{disgust} (DI), \enquote{happy} (HA) and \enquote{neutral} (NE) the system is able to recognize them with 91\% classification accuracy rate. \enquote{Fear} (FE) got 95\% recognition rate. The final average recognition rate is 94.85\%. 

\begin{table}[]
\centering
\caption{Classification Accuracies (\%) on the JAFFE Database.}
\label{table3}
\begin{tabular}{|c|c|}
\hline
{\color[HTML]{000000} \textbf{Approach}} & {\color[HTML]{000000} \textbf{Average Recognition Rate \%}} \\ \hline
{\color[HTML]{000000} SFER (ours)} & {\color[HTML]{000000} 94.85} \\ \hline
{\color[HTML]{000000} LC-KSVD-1} & {\color[HTML]{000000} 76} \\ \hline
{\color[HTML]{000000} LC-KSVD-2} & {\color[HTML]{000000} 78} \\ \hline
{\color[HTML]{000000} CAE-based} & {\color[HTML]{000000} 95.8} \\ \hline
{\color[HTML]{000000} FIS} & {\color[HTML]{000000} 87.6} \\ \hline
{\color[HTML]{000000} Sobel-based} & {\color[HTML]{000000} 93.1} \\ \hline
\end{tabular}
\end{table}

We compare our approach with other sparse approaches LC-KSVD1 and LC-KSVD2 \cite{29} but also with other techniques: Convolutional Autoencoder, Sobel-Based, Fuzzy Inference System \cite{hamester2015face}. Table \ref{table3}, shows the average recognition rate for those different approaches. It is obvious that our approach outperforms other sparse approaches and exhibits performance similar to those of the most recent state of the art methods.

\subsection{Model Evaluation over the DynEmo Database}
\label{evaluation}

\begin{figure}[]
\centering
\includegraphics[width=3 in]{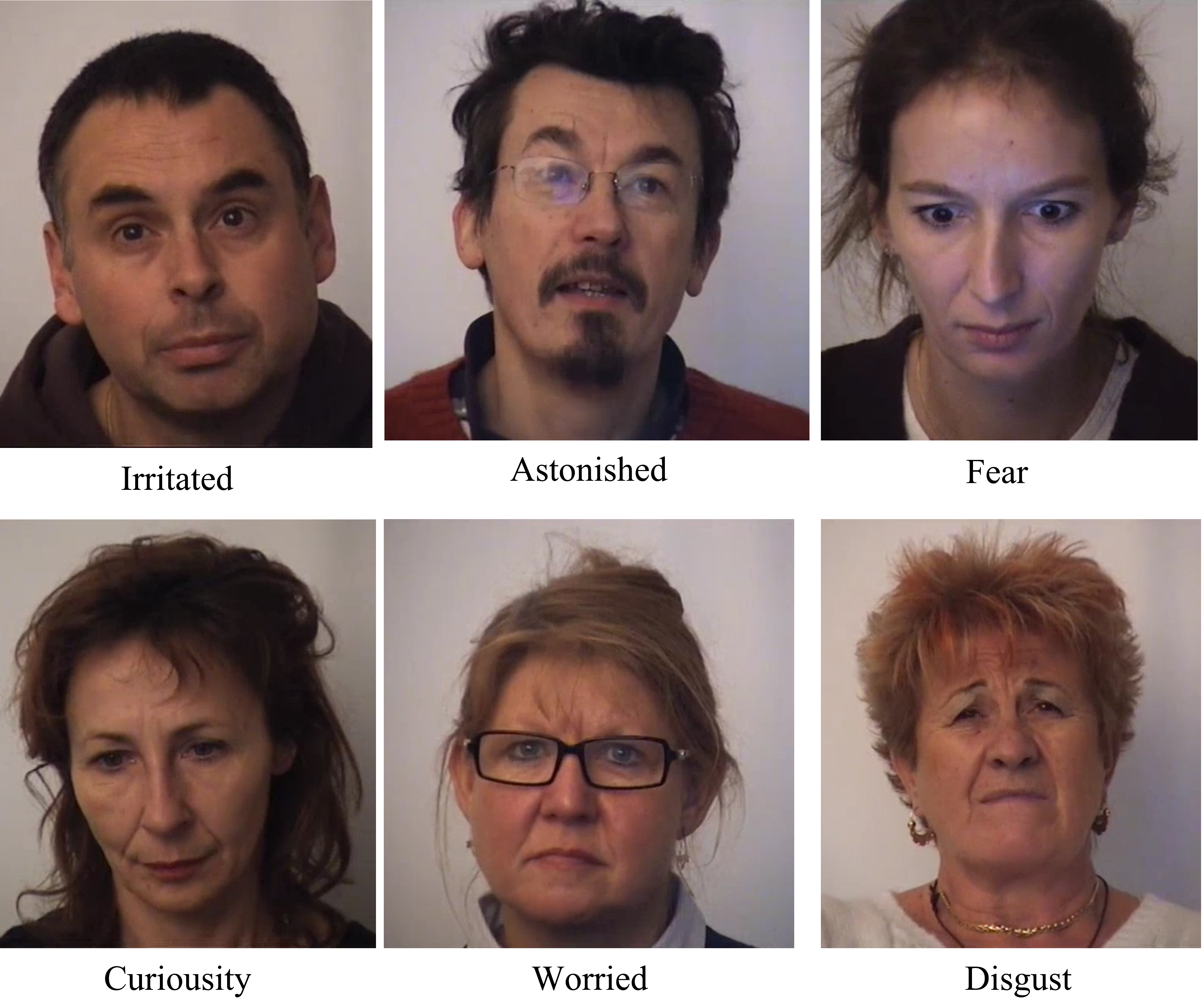}
\caption{Examples of spontaneous facial expressions from the DynEmo database.}
\label{fig:DynEmo}
\end{figure}

\begin{figure}[]
\centering
\includegraphics[width=3 in]{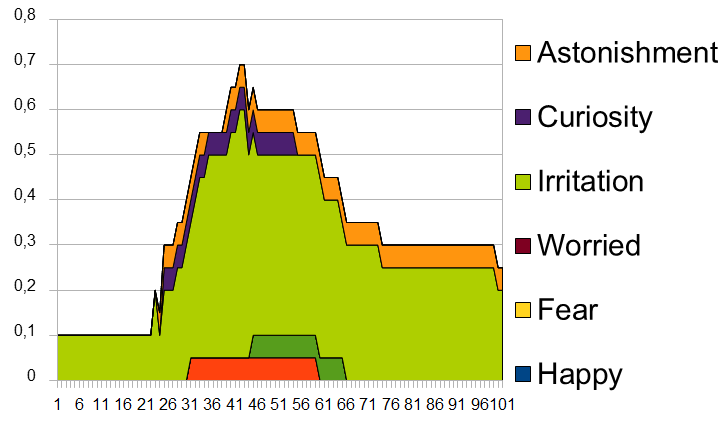}
\caption{Time line of continuous annotations.}
\label{fig:EFERecord}
\end{figure}

\textbf{The DynEmo Database:}
\noindent DynEmo is a database containing dynamic and natural emotional facial expressions (EFEs). It is made of six spontaneous expressions which are: \enquote{irritation}, \enquote{curiosity}, \enquote{happiness}, \enquote{worried}, \enquote{astonishment}, and \enquote{fear} (see figure \ref{fig:DynEmo}). Those expressions have been elicited by showing some emotive short clips to volunteer subjects. The database contains a set of 125 recordings of EFE of ordinary Caucasian people (ages 25 to 65, 182 females and 176 males) filmed in natural but standardized conditions. In this set, EFE recordings are both associated with the affective state of the expresser itself and with continuous annotations of observers' ratings of the emotions displayed throughout the recording (see figure \ref{fig:EFERecord}). The x-axis of figure \ref{fig:EFERecord} represents the time line, while the y-axis represents the probability of judgment for each frame. In the rest of this paper, we will refer to the expresser as encoder and to the observer as decoder.  Figure \ref{fig:EFERecord} shows that $10\%$ of the decoders recognized the feeling of the encoder as irritation at the beginning of the video (time line: frame 1 to frame 20). While for the time line between frame 21 and frame 76, the judgement of different decoders led to different results. To overcome this problem, when different decoders judge with different classes, we associate to each frame the class that gets the highest probability. For example, for the time line corresponding to frames between 36 and 41, the apex (maximum expressiveness of the emotion) is associated to the astonishment class with a probability of $70\%$. In some cases, when the probability of two or more different classes is the same, we refer to the previous frame judgment as additional information to judge the current frame.

\begin{figure}[!h]
\centering
\includegraphics[width=3 in]{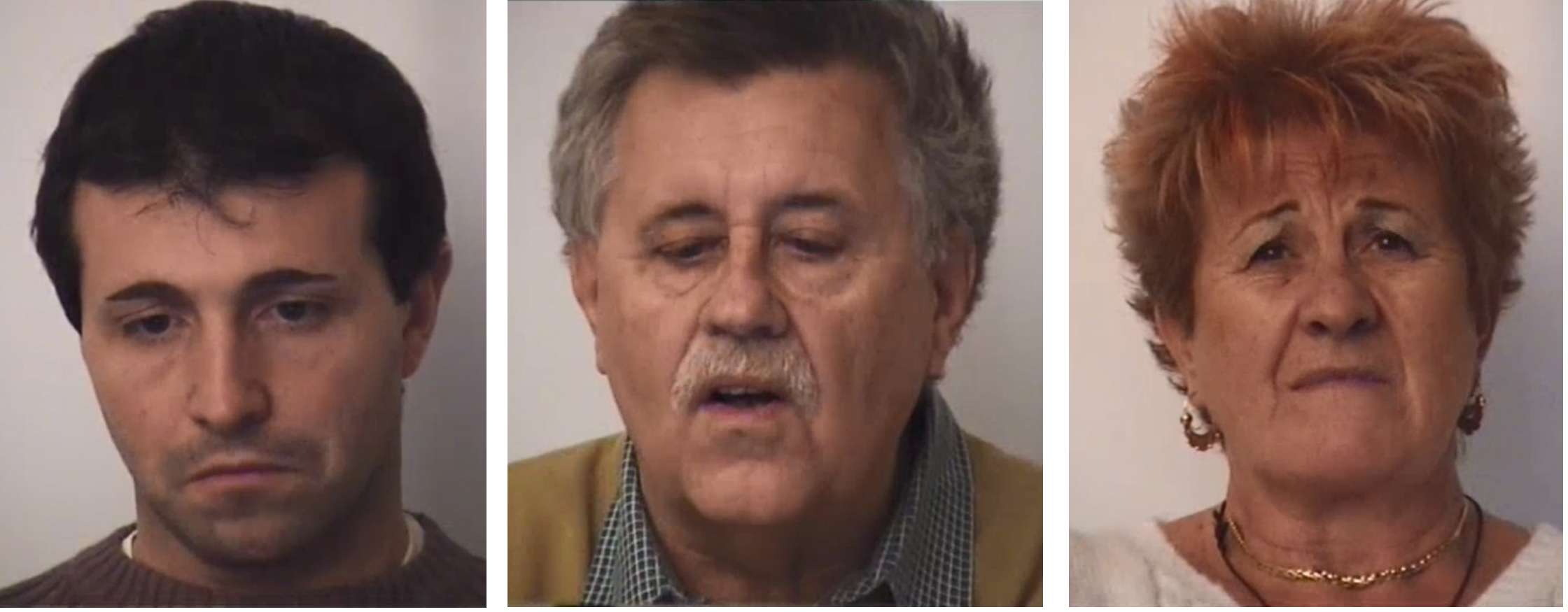}
\caption{Same spontaneous emotion (disgust) expressed by different encoders.}
\label{fig:DIS}
\end{figure}

We built a labelled spontaneous database in which the frames are extracted based on the previous defined ground truth. 480 images of female and male facial expressions of 65 different identities form the database. Each image has a resolution of $239 \times 200$ pixels after face detection and cropping. The head is not totally in frontal pose. The number of images corresponding to each of the six categories of expressions is roughly the same (80 images per class). The dataset is challenging since it is closer to natural human behaviour and figure \ref{fig:DIS} shows that even for the same emotion, people can perform it in different ways.

\vspace{3mm}
\noindent \textbf{DynEmo Database Protocol:}
Identities that appear in the training data sets do not appear in the test set.
\begin{enumerate}
     \item \textbf{train set}: 60 images per class are picked out as training set. In total we have 360 facial expression images, randomly shuffled, for training our algorithm.
     \item \textbf{development set}: Leave-one-out cross validation is considered over the training set to tune the algorithm parameters.
     \item \textbf{test set}: 20 images per class are picked out as test set. In total we have 120 facial expression images, randomly shuffled, to test the performance of our algorithm.
\end{enumerate}

\vfill
\textbf{Experimental Setup and Analysis: }
\vfill

Same experimental setup as in the control experiment has been followed. \textbf{Firstly}, the dataset is divided into two portions based on the number of images per class as defined in the previous DynEmo dataset protocol. Then, RFFD is applied in which the optimal random matrix that generates good discriminative features and the optimal dimensionality size ($n$-features = 250 feature points) are derived. Therefore, passing through the first stage of SFER algorithm, we get a training set with $360$ images and $250$ feature points ($n$) in addition to a testing set of $120$ images and $250$ feature points. The average recognition rate over the projected data prior to the second stage of SFER algorithm is $68.4\%$. The performance of RFFD  is compared with PCA which achieves an average recognition rate of $24\%$ only. It is obvious that PCA is not powerful at all to extract good discriminative features compared to RFFD when considering spontaneous facial expressions.

\textbf{Secondly}, a dictionary of size ($250$ features, $360$ atoms ($K$)) is initialized (projected training set). The sparsity level ($L$) is estimated to $10\%$ of the dictionary size by controlling the absolute reconstruction error. The dictionary is refined via K-SVD and the sparse matrix is derived via OMP algorithm.

\textbf{Finally}, a linear SVM classifier is trained over the training matrix ($360$, $360$). The test sparsity matrix ($120$, $360$) is used to assess the ability of the classifier for generalization. A grid search is applied to find the best regularization parameter $C$, where $C$ is found to be $10$.

\begin{table}[]
\centering
\caption{Confusion Matrix and average recognition rate per class (RR) in $\%$ .}
\label{tabelconf}
\begin{tabular}{|c|c|c|c|c|c|c|c|}
\hline
\textbf{Class} & IRR & CU & HA & WO & AST & FE & \textbf{RR} \\ \hline
IRR            & 85  & 5  & 5  & 0  & 5   & 0  & \textbf{81}                                  \\ \hline
CU             & 5   & 95 & 0  & 0  & 0   & 0  & \textbf{95}                                \\ \hline
DI             & 0   & 0  & 95 & 5  & 0   & 0  & \textbf{93}                                  \\ \hline
WO             & 15  & 0  & 5  & 80 & 0   & 0  & \textbf{86}                                  \\ \hline
AST            & 0   & 0  & 0  & 0  & 100 & 0  & \textbf{98 }                                 \\ \hline
FE             & 5   & 0  & 0  & 0  & 0   & 95 & \textbf{97 }                                 \\ \hline
\end{tabular}
\end{table}

Table \ref{tabelconf} shows the confusion matrix and the average recognition rate per class. It appears that the highest number of misclassifications is obtained for \enquote{irritation} (IRR) and \enquote{worried} (WO). Figure \ref{fig:DynEmo} shows that WO and IRR expressions are visually close to each other. For the rest of the classes like \enquote{curious} (CU), \enquote{astonishment} (AST) and \enquote{fear} (FE), the obtained recognition rate is above 95\%. The class \enquote{disgust} (DI) got a 93\% recognition rate. The average recognition rate is 91.67\%. Table \ref{table6} shows the recognition rate on the DynEmo dataset compared to the other sparse approaches. It can be seen that our approach performs much better than LC-K-SVD1 and LC-K-SVD2.

\begin{table}[]
\centering
\caption{Classification Accuracies (\%) on the DynEmo Database.}
\label{table6}
\begin{tabular}{|c|c|}
\hline
{\color[HTML]{000000} \textbf{Approach}} & {\color[HTML]{000000} \textbf{Average Recognition Rate \%}} \\ \hline
{\color[HTML]{000000} SFER (ours)} & {\color[HTML]{000000} 91.68} \\ \hline
{\color[HTML]{000000} LC-KSVD-1} & {\color[HTML]{000000} 20.1} \\ \hline
{\color[HTML]{000000} LC-KSVD-2} & {\color[HTML]{000000} 85.4} \\ \hline
\end{tabular}
\end{table}

\begin{table*}[]
\centering
\caption{Average recognition rate using SFER approach}
\label{combination}
\begin{tabular}{|c|c|c|c|c|c|c|c|c|c|c|}
\hline
\textbf{\begin{tabular}[c]{@{}c@{}}Number of\\ images per class\end{tabular}}                & \textbf{AN} & \textbf{IRR} & \textbf{SU} & \textbf{CU} & \textbf{SA} & \textbf{AST} & \textbf{HA} & \textbf{WO} & \textbf{DI} & \textbf{FE} \\ \hline
\textbf{Training set}                                                                        & 20          & 60           & 20          & 60          & 20          & 60           & 20          & 60          & 80          & 80          \\ \hline
\textbf{Test set}                                                                            & 10          & 20           & 10          & 20          & 10          & 20           & 10          & 20          & 30          & 30          \\ \hline
\textbf{\begin{tabular}[c]{@{}c@{}}Average recognition \\ Rate per class in \%\end{tabular}} & 93          & 78           & 94          & 90          & 88          & 92           & 89          & 83          & 89          & 88          \\ \hline
\textbf{\begin{tabular}[c]{@{}c@{}}Final average\\ recognition rate\end{tabular}}            & \multicolumn{10}{c|}{88.4 \%}                                                                                                               \\ \hline
\end{tabular}
\end{table*}

\subsection{Generalization Performance}

The system's generalization performance is evaluated on the combination of the two datasets: JAFFE + DynEmo. Table \ref{combination} show the distribution of the new database, the average recognition per class and the final average recognition rate obtained over the new database. Same experimental setup as the two previous experiments has been followed. The model is tuned by performing 10-folds cross validation over the training set and tested over the test set. Table \ref{combination} shows that our model is capable of recognizing different classes related to different emotions and mental states.
88.4 \% as an average recognition rate over the 10 classes is achieved.

\section{\uppercase{Conclusion}}
\noindent In this paper, a robust spontaneous facial expression recognition algorithm (SFER) based on facial images that recognizes non-basic affective state including mental state is presented. We developed a method to pre-train the dictionary that enforces sparsity and enhances dictionary performance. We shown that it was possible to learn under-complete dictionary once good discriminative features are extracted prior to dictionary refining stage which ensures the uniqueness of the selected atoms from the dictionary during the optimization process. We proposed the use of random projection as a mean of dimensionality reduction and as a mean of solving the problem of shared subspace. We exhibited very good recognition rates over the recent spontaneous facial database DynEmo. A possible work for the future is exploiting the temporal dynamics of facial expressions in order to improve the recognition rates. Temporal information might be useful since expressions not only vary in their facial deformations but also in their onset, apex, and offset timings. 

\vfill
\bibliographystyle{apalike}

\vfill
\end{document}